# A Hybrid Multi-Objective Carpool Route Optimization Technique using Genetic Algorithm and A* Algorithm

Romit S Beed[a], Sunita Sarkar[b], Arindam Roy[c], Suvranil D Biswas[a], Suhana Biswas[a]

[a] Department of Computer Sc., St. Xavier's College (Autonomous), Kolkata
[b] Department of Computer Sc. & Engineering, Assam University, Silchar,
[c] Department of Computer Sc., Assam University, Silchar

Keywords: Carpooling, Genetic Algorithm, A* Algorithm, Pareto Optimality, Path-finding

**A b s t r a c t**

Carpooling has gained considerable importance in developed as well as in developing countries as an effective solution for controlling vehicular pollution, both sound and air. As carpooling decreases the number of vehicles used by commuters, it results in multiple benefits like mitigation of traffic and congestion on the roads, reduced demand for parking facilities, lesser energy or fuel consumption and most importantly, reduction in carbon emission, thus improving the quality of life in cities. This work presents a hybrid GA-A* algorithm to obtain optimal routes for the carpooling problem in the domain of multi-objective optimization having multiple conflicting objectives. Though Genetic algorithm provides optimal solutions, A* algorithm because of its efficiency in providing the shortest route between any two points based on heuristics, enhances the optimal routes obtained using Genetic algorithm. The refined routes, obtained using the GA-A* algorithm, are further subjected to dominance test to obtain non-dominating solutions based on Pareto-Optimality. The routes obtained maximize the profit of the service provider by minimizing the travel and detour distance as well as pick-up/drop costs while maximizing the utilization of the car. The proposed algorithm has been implemented over the Salt Lake area of Kolkata. Route distance and detour distance for the optimal routes obtained using the proposed algorithm are consistently lesser for the same number of passengers when compared with the corresponding data obtained using the existing algorithm. Various statistical analyses like boxplots have also confirmed that the proposed algorithm regularly performed better than the existing algorithm using only Genetic Algorithm.

## 1. Introduction

In the twenty-first century, due to the constant development of society and industry, the need for mobility has increased rapidly and so has the use of cars, especially in the developing and under-developed countries [1]. Affording cars presently is within the reach of many due to various bank schemes and loans, but it inevitably results in other issues like elevated pollution levels, traffic jams and increased monthly expenditure of the individuals. According to Environment Canada, air pollution by vehicular emissions results in health problems like cardiovascular and respiratory diseases, various allergies along with some neurological effects. Carpooling is an effective solution to these major problems [2,3]. Research also suggests that carpooling results in less stress than commuting alone. Carpooling is the most rapidly evolving solution for the shift from vehicle ownership to shared vehicle usage mobility. The future of mobility consists of technology-enabled, door-to-door, multi-modal travel encompassing pre-trip, in-trip, and post-trip services to improve journey experience to the Mobility User [4]. Carpooling helps users to share a ride to destinations in the same area, by either casual carpooling or by real-time carpooling. According to the most commonly used terminology, carpooling is the agreement of sharing the use of a particular car by many passengers, usually commuting along the same route/ journey at mutually compatible times [5]. As carpooling decreases the number of vehicles used by travelers, it results in various benefits like mitigation of traffic or congestion on the roads, reduced demand for parking facilities, lesser energy or fuel consumption, and most importantly, reduction in carbon emission, thus improving the quality of life in cities.

Genetic Algorithms (GA) represent a class of optimized, adaptive, and iterative algorithms that function upon existing data sets and design the developing concepts on the basis of genetic information as observed in nature. GA operates on a set of solutions using operations like mutation, crossover and selection and stops only when concurrency of the required criterion takes place. These algorithms, even though irregular or randomized, mostly use heuristic data to utilize promising regions within the search space [6]. A population in a GA is a set of various coinciding search points or solutions. A new population is produced for each iterative step, called a generation. A solution, often labeled as a chromosome, $y = [y1, .., yn]$, is basically a set of variables in a search space of n dimensions or variables [7]. These n variables are similar to n-genes. This paper utilizes the essence of GAs to provide multiple solutions that are non-dominated, that is to say, equally important when considering a broad array of conflicting objectives. Any of these non-dominated solutions can be used as the final route without having the driver or the passengers suffer a loss in their interests [8].

One of the most extensively used path-finding algorithms is the A* algorithm, which is a heuristic or an informed search algorithm. A* uses the fundamental concepts of the Greedy Best First Search technique with the Dijkstra Algorithm and provides the shortest accessible path between the source and the destination. A* algorithm is used majorly in the fields of game development, robotics, traversal of graphs and maps, etc. The most important features of this algorithm are its high efficiency and its convenience. It uses a valuation function, denoted by $f(n)$, as a guiding capacity to find the required path, both effectively and precisely. This valuation function that gives an estimated cost of the path from the starting node to the target node, via the intermediate node is expressed as

$$f(n)= g(n) + h(n) \qquad (1)$$

where g(n) and h(n) are the actual cost from the starting node to the current node(n) and the assessed/heuristic cost of the shortest path from this current node(n) to the target node, respectively. The heuristic cost function, h(n), for each intermediate node is calculated by taking the Euclidean distance between the current node(n) and the target node. h*(n) denotes the actual cost of the best possible path from the present node(n) to the objective node. Now, if h*(n) ≥ h(n) for all intermediate nodes, then it is accepted as a reachable route-finding process. It has vast uses in the fields of GIS systems along with game routing systems and maps.

There are numerous pieces of existing literature that have proposed algorithms for the carpooling problem. Some authors used Genetic Algorithm, weighted sum methodology, pathfinding algorithms like A* algorithm, Dijkstra algorithm, etc. Unfortunately, there are not many optimized car-pooling algorithms that support the users along with the service providers to choose the most optimal routes, keeping in mind the various real-life constraints that affect this decision-making process. This work proposes a hybrid algorithm that implements carpool route optimization using a Genetic Algorithm and refines the route using the A* algorithm. This work aims at providing a choice of optimal routes, to facilitate the passengers and the service provider/driver by maximizing car utilization, minimizing total distance travelled, as well as keeping in consideration the individual passenger's cost. Instead of using the A* algorithm in its traditional form, where the parameter used to select the most optimal route is the distance, here the A* algorithm is used to optimize the set of routes obtained from the genetic algorithm. Considering the aforementioned conflicting objectives, viz. maximizing car utilization, minimizing total distance travelled, minimizing individual passenger's cost, the authors provide a list of non-dominated routes which are all considered equally good for both the passengers and the service provider.

The primary objective for this work is to provide a set of non-dominated routes to solve the traditional carpooling problem. It has been motivated by the fact that carpooling is very essential in the present scenario and is one of the most effective means in dealing with the detrimental effects of pollution. Carpooling works towards providing a greener environment by encouraging riders to share rides. Not only does this technique greatly reduces the fuel consumption per person and benefits the environment, it also provides a cost-effective mode of travel for the riders by allowing them to travel together and share the cost leading to financial savings. Another objective of this work is to deal with the problem of deviating from the main route to pick-up/drop passengers and returning back to that point before continuing the journey. This leads to excess and redundant travel. This motivated the authors to integrate the A* algorithm with GA to produce optimized routes with greater efficiency by fetching shorter new routes from the pick-up/drop point to the final destination.

## 2. Literature Survey

Varied research works have been published in the arena of carpooling to find optimal routes as well as to allocate riders while matching their requirements. The concept of carpooling consisted of picking up passengers in sequence and dropping them later. It later developed into a "park and ride" concept having a common pick up point for all the riders. Recently, due to the growth and spread of the internet technology, dynamic carpooling, i.e. picking up and dropping off passengers as and when requests arise while travelling, has reached its peak. This modern version of car-pooling witnessed its first practical use when John Zimmer, from Cornell University along with Logan Green, from University of California, created "Zimride", a dynamic match-making service to connect drivers and passengers using GPS on android phones [9].

Knapen et al. [10] developed a model to coordinate ridesharing trips. Clients registered their profile and periodical information about repeating trips, and the service provider prompted the enlisted users to combine their outings through ridesharing. The service provider assessed the satisfaction quotient of co-travelers based on prior information. Another model was developed by Schreieck et al., [11] which focused on matching ride-sharing offers with ride requests and also storing and retrieving routes using inverted index data structures. Google API was used for geocoding the source and destination address. This system employed the matching mechanism by emphasizing various shortest path algorithms, such as Dijkstra's Algorithm and A* Search. It was observed that the proposed technique performed well enough for real-time applications while being simpler than existing optimization-based techniques. Another model by He et al. [12] concentrated on the profitability of the ride. Different GPS directions were mined to get the frequently utilized routes using route parting and gathering, grid mapping techniques, etc. An improved carpool system was developed by Karande et al. [13] that allowed users to avail the services of ridesharing via a smartphone. This model defined an advanced Genetic algorithm based carpool route and matching algorithm that provided a solution by securing ideal match arrangements.

Masum et al. proposed an innovative model [14] to solve the carpooling problem that used a fitness function to select desirable parents to reproduce and create the next generation. The process of preparation ensured the removal of duplicate genes within the child as a result of crossover and mutation. Missing genetic information was re-inserted using a heuristic method. Another work by Rathod et al. [15] proposed a GA based carpooling service that generated optimal routes of travel within a short period of time. The proposed algorithm generated intermediate paths that were used to find the solution to the empty seats available in the car. Later another model was developed by Boukhater et al. [16] that was map-based and provided shared rides for all customers, considering their personal inclinations. The proposed algorithm performed better than the traditional algorithm. A heuristic algorithm for Maximum Carpool Matching was proposed in a paper by Hartman et al. [17] and it demonstrated the Maximum Carpool Matching problem was NP-hard even for the situation where the weight function is binary. They [18] introduced a natural integer linear program and demonstrated that if the arrangement of drivers is known, an optimal assignment of travelers to drivers can be found in polynomial time utilizing a reduction to Network Flow.

A* Search is a procedure majorly utilized in the field of Artificial Intelligence. A model was proposed by Sharma et al. [19] that used a bi-directional search technique on the traditional A* algorithm for finding the shortest path. As, A* algorithm is in general one of the most optimal path-searching algorithms that use heuristics, optimizing it even further by applying the bi-directional search, resulted in a system that provided the shortest possible path, in very less search time. The A* algorithm, both in its unidirectional and its bi-directional forms, provided results much better than those of the Dijkstra algorithm in its traditional and bi-directional forms, respectively. The authors concluded that the A* algorithm outdid the Dijkstra algorithm in all informed search situations, with and without obstacles. Another innovative work by Arnates et al. [20] proposed a hybrid algorithm consisting of a heuristic approach applied on a Genetic Algorithm to provide facilities of path-finding and re-routing in cases of Unmanned Aerial Vehicles (UAVs). The algorithm here involves greedy heuristic to find possible paths and then uses the GA to provide the most optimal solutions within a comparatively low amount of time. To prove the efficiency of the proposed algorithm, experimental simulations were conducted, results of which showed that this combination of the given heuristic approach with GAs is a good strategy for routing UAVs.

Zeng et al. [21] performed various tests on road maps of two regions of California to compare the effectiveness of the shortest path algorithms of the Gallo-Pallottino (GP) class with the A* algorithm and its three variations. The authors successfully proved that the A* algorithm and its variations performed much faster and better compared to what the GP-class algorithms do on real-life road maps or networks. It also showed that on-road networks, A* algorithm's performance exceeds even the most optimal execution of the Dijkstra Algorithm, that too by a very large margin. This work proved that the A* algorithm's optimality increases with the increase in the size of the road networks. The experiments in this research work also showed that one of the three variations of A*, A star with approximate buckets (ASBA) outperformed all the other algorithms that were taken into account.

A variation of the native A* algorithm, called the A* Hamilton algorithm by Halaoui et al. [22] was proposed to navigate to many destinations in any order. The algorithm provided the shortest path from a source location to many destinations without any particular order. Meng et al. [23] proposed another model where a salient feature of the A* algorithm was to move towards the direction of the destination by utilizing directional elements, with the goal that the intermediate route procedure will move towards the shortest path as soon as possible. Then, the direction factor was utilized to guarantee that the priority of the path finding of the A* algorithm was to move towards the direction of the target. The proposed technique improved the efficiency of the algorithm as the outcome of the A* optimization algorithm was around 20–50% better than the traditional A* algorithm. Also, the best case was achieved at around 89%. Herbawi et al. [24] proposed a model that dealt with the car-pooling problem in the form of a dynamic and multi-objective ride-sharing situation. This model assigned passengers to the car drivers and characterized the user requests and coordinated the passengers' pickup and drop off timings, optimally. They proposed a hybrid algorithm that acted at two levels and divided each day into a group of time periods, to deal with the ride-matching problem using time windows. A hybrid path-finding model using Genetic Algorithm was proposed by Yui et al [25]. It provided a multi-weighted heuristic (MWH) function, which was then used in the A* algorithm to find the most optimal routes. GA provided multiple heuristic functions that acted as agents, which in turn competed with each other to produce children chromosomes or agents. On optimizing all these agents, the final MWH function was returned.

The authors of this paper, inspired by the aforementioned researches, thus proposes a hybrid model that uses a genetic algorithm for route optimization and aims to use A*algorithm to refine the proposed optimal solution for the carpooling problem. The hybrid route search technique controls the search towards the destination node by using lower limits on the distance to the target. The proficiency of this approach relies on the lower values. A* search utilizes path costs along with heuristic values. Here, along with the lower bounds on the distance to the target, the authors optimize the traditional A* algorithm by incorporating other parameters, like the density of ride requests generated in a route, the total detour taken for picking up and dropping off passengers and the length of the route, to provide a result that is optimal for both the passengers and the driver.

## 3. Proposed Carpooling model

The proposed hybrid carpooling algorithm attempts to solve a real-world problem from the domain of multi-objective optimization consisting of conflicting objectives. The model proposes to obtain an optimized route that would generate the maximum profit for the service provider. A detailed analysis of the car routing problem brings into light the presence of multiple conflicting objectives. The essence of multi-objective optimization is that it strives to obtain a set of solutions, by applying various mechanisms, so that no particular objective is neglected on the behest of others. This work aims to improve the Hierarchical Multi-objective Route Optimization for Solving Carpooling Problem by Beed et al. [24] which used the concepts of Genetic Algorithms and proposed a carpooling model that took into consideration multiple conflicting objectives. While solving a carpooling problem, the basic intuition is frequently directed towards minimizing the distance traveled by the passengers. As shown in the existing work [26], taking other factors such as occupancy, detour, and total cost into consideration can significantly provide better selection of a route. The proposed model there solved the car-pooling problem by dividing these multiple objectives into a hierarchical model, in order to optimize the solution. Since the higher-level objectives do not completely control the parameters of the lower level, the lower level of this hierarchy consists of the discording objectives that are confined to the individual passengers only, namely detour distance for pickup and drop of a passenger and passenger density of the surrounding area. The higher-level objectives consisted of minimizing the distance traveled by the car, maximizing the utilization of the vehicle and reducing the pick-up and drop-off cost for a particular passenger. Hierarchical decision structures [27] help in realizing real-life situations better. This work provided a route that was optimal for the users and also profitable for the service provider [28].

The proposed hybrid algorithm aims to improve the existing algorithm by including A* algorithm. Since the A* algorithm is efficient in providing the shortest route between any two points, the authors have utilized this characteristic to generate a route incorporating all the pick-up and drop-off points of passengers for every corresponding elite route. This leads to maintaining the same level of occupancy while benefiting greatly in other aspects of total distance and detour. Consider Passenger A needs to travel from point X to point Y in Figure 1. There are three available paths from X to Y for the passenger to travel through. The routes have been colored Red, Blue, and Purple. The Red path gives the shortest distance between the two points and has 4 probable passenger requests along this route only. The Purple path is comparably longer than the Red path but has the highest number of requests along the route. Now, the Blue path has fewer requests along the route than Purple, but the detour distances for each of these requests is quite less as all the passengers are located close to the main route. All the routes are equally favorable as they are superior with respect to a certain objective.

Once the optimal routes are generated by one generation of GA, the routes are now processed using the A* algorithm to obtain better solutions. As can be perceived from Fig.1, the routes generated by the GA require the cars to take additional detours from the route to pick-up/drop passengers and return back to the route. This leads to an unnecessary increase in travel and can be a drawback in cases where it is much more intuitive to travel directly to the next pickup/drop location. The routes that have been optimized by the GA-A* algorithm incorporates the pick-up/drop points thus eliminating the need for additional detour. The algorithm also ensures that the car travels through the shortest path between successive pick-up/drop points. Thus, a shorter route is generated by using the hybrid algorithm using A* algorithm. These routes are further subjected to dominance tests to obtain Pareto Optimal solutions. The hybrid algorithm is as follows:

ALGORITHM(GA-A*):
Step 1:     Read passenger request log
Step 2:     If request log is empty then go to Step 18
Step 3:     If the passenger pickup point is within a radius of **t** kms
            then    store pickup and drop location of passenger **i** into **X** and **Y**
                    go to step 4

|          |                                                                                         |
|----------|-----------------------------------------------------------------------------------------|
| | else    go to step 1 |
| Step 4:  | Set **X** and **Y** as source and destination of route; gen =1                          |
| Step 5:  | Generate randomly a pool of **m** chromosomes being routes between **X** and **Y**.     |
| Step 6:  | Select randomly **q** chromosomes for first generation of Genetic Algorithm             |
| Step 7:  | For each generation of GA perform Tournament Selection, Crossover and Mutation          |
| Step 8:  | Perform Dominance test to obtain Pareto Optimal **n** chromosomes / optimal routes.     |
| Step 9:  | For each route obtained in Step 8, do                                                   |
| Step 10: |     For each point **P** (pickup/drop point) on the route, do       |
| Step 11: |         Use A* algorithm to obtain shortest path from **P** to **Y**. |
| Step 12: |         Append path **X** to **P** with this new shortest path from **P** to **Y**. |
| Step 12: |         Implement dominance tests on the newly generated route to maintain Pareto Optimality. |
| Step 13: |     End of Step 10 Loop                                             |
| Step 14: | End of Step 9 Loop                                                                      |
| Step 15: | Combine **n** Pareto optimal chromosomes with another set of **k** random chromosomes from the pool for a total of **q** chromosomes. |
| Step 16: | gen = gen + 1                                                                           |
| Step 17: | Go to Step 7 till gen < max_gen                                                         |
| Step 18: | Print **n** Pareto optimal solutions                                                    |
| Step 19: | Exit                                                                                    |

The A* algorithm functions in the following manner:

| | |
|---|---|
| Step 1: | Insert starting node into Open List (OL) |
| Step 2: | Retrieve the first node of OL as the current node (CN) |
| Step 3: | If CN is the destination node, then exit |
| Step 4: | Explore the neighboring nodes of CN |
| Step 5: | Set CN as their parent and calculate valuation functions |
| Step 6: | If neighbors of CN are not present in Closed List (CL) |
| Step 7: |     Insert neighbors of CN in OL in increasing order of valuation functions. |
| Step 8: | Remove CN from OL and add to CL. |
| Step 9: | Go to Step 2. |

## 4. Experimentation and Results

The map of Salt Lake area of Kolkata has been used as a prototype to implement this algorithm. The map was divided into 116 junction points or nodes. The Google Map API was used to obtain the actual distance between the nodes and a corresponding distance matrix was created. A request matrix was dynamically generated and used by the algorithm to generate routes. The initial request was randomly generated.

On executing the program code, the following results are obtained.

| GA Route | GA-A* Route | Occupancy | Total Dist. GA | Total Dist. GA-A* | Detour GA | Detour GA-A* | Percentage Improvement in Distance |
|---|---|---|---|---|---|---|---|
| 73\|84\|92\|93\|105\|106\|107\|115\| | 73\|84\|92\|93\|105\|106\|112\|106\|103\|106\|112\|115\| | 5 | 7280 | 4320 | 4300 | 1400 | 40.66% |
| 73\|84\|92\|93\|94\|93\|105\|106\|112\|115\| | 73\|84\|92\|93\|105\|102\|94\|95\|94\|93\|105\|106\|112\|106\|103\|106\|112\|115\| | 6 | 9120 | 6220 | 5600 | 3300 | 31.80% |
| 73\|60\|73\|84\|92\|93\|105\|106\|107\|115\|\| | 73\|84\|92\|93\|105\|106\|112\|106\|103\|106\|112\|115\| | 5 | 7880 | 4320 | 4300 | 1400 | 45.18% |
| 73\|84\|92\|108\|109\|112\|115\| | 73\|84\|92\|93\|105\|106\|112\|106\|105\|106\|112\|115\| | 4 | 8020 | 4420 | 4400 | 1500 | 44.89% |
| 73\|84\|92\|108\|109\|112\|109\|112\|109\|112\|115\| | 73\|84\|92\|93\|105\|106\|112\|106\|105\|106\|112\|115\| | 5 | 11420 | 4420 | 6000 | 1500 | 61.30% |
| 73\|84\|92\|93\|105\|102\|94\|102\|103\|104\|107\|106\|105\|109\|112\|115\| | 73\|84\|92\|93\|105\|102\|103\|102\|94\|95\|94\|102\|105\|102\|103\|97\|103\|106\|107\|106\|112\|106\|103\|106\|105\|109\|112\|115\| | 10 | 15760 | 9680 | 10060 | 6760 | 38.58% |
| 73\|74\|61\|74\|73\|84\|92\|108\|109\|112\|115\| | 73\|74\|61\|74\|86\|87\|95\|94\|93\|105\|106\|112\|115\| | 4 | 9120 | 3880 | 3600 | 960 | 57.46% |
| 73\|74\|73\|84\|92\|93\|94\|85\|94\|93\|105\|109\|112\|106\|107\|115\| | 73\|84\|92\|93\|105\|102\|94\|95\|87\|95\|94\|93\|105\|109\|112\|106\|105\|102\|103\|106\|112\|115\| | 7 | 14230 | 7540 | 7900 | 4620 | 47.01% |
| 73\|84\|92\|84\|92\|108\|92\|108\|109\|112\|115\| | 73\|84\|92\|93\|105\|109\|105\|93\|92\|93\|105\|106\|112\|106\|105\|106\|112\|115\| | 5 | 12620 | 6720 | 5800 | 3800 | 46.75% |
| 73\|84\|92\|84\|73\|74\|86\|88\|86\|88\|89\|90\|91\|97\|103\|106\|107\|106\|112\|115\|\| | 73\|84\|92\|93\|92\|84\|73\|74\|75\|76\|77\|78\|79\|80\|98\|104\|103\|106\|105\|102\|103\|106\|112\|106\|105\|106\|105\|109\|112\|115\| | 9 | 17640 | 11570 | 10560 | 8650 | 34.41% |

**Table 1: The data displayed in the above table represents the result of a particular execution**

Table 1 allows a comparative study of the routes provided by the GA and the GA-A* algorithm, their corresponding detours and total distance. The distances have all been calculated in meters. Although the routes under the 1$^{st}$ column appears to contain a fewer number of nodes and hence, by intuition, should have a lesser total distance, it is to be noted that these routes just represent the basic travel path of the car and does not take into account the detours for picking up and dropping off passengers, which need to be calculated separately. Since the routes generated by the GA-A* algorithm includes the pickup and drop-off points of the passengers, there is no additional detour. For a comparative study of the detour between GA-A* algorithm and the GA algorithm, it is assumed that the car deviates from the shortest path between the start and end points of the request only for the sake of picking up passengers, and hence, the detour has been calculated as the difference between the total distance of the route and the total distance between the shortest path between the start and end points of the request.

The following chart (Figure 2) has been constructed using the data summarized in the last column of the above table 1 to visualize the improvement in the total distance of the routes over a single execution, after being optimized using the GA-A* algorithm. The improvement is calculated as a percentage of the decrease in total route distance over the total distance of the routes generated by the GA.

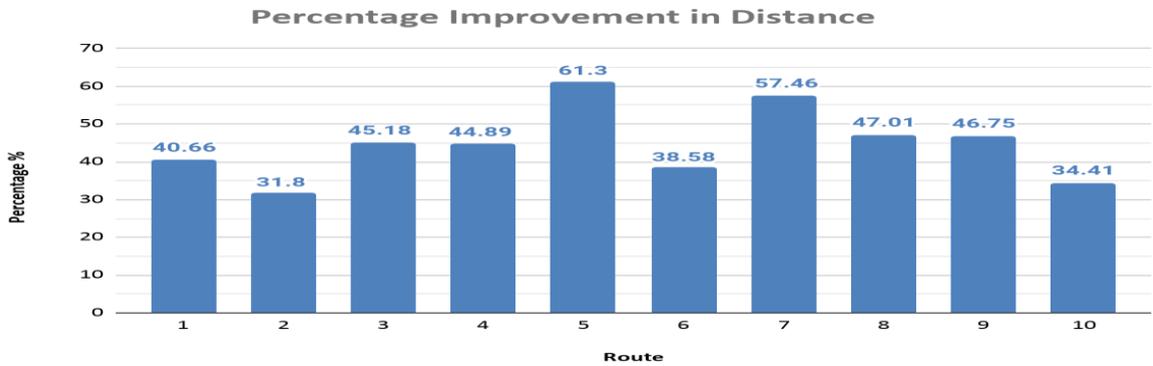

Fig. 2: Percentage improvement in Distance over a single execution

**Statistical Analysis for comparison of Results obtained by Hybrid GA-A* algorithm and existing GA.**

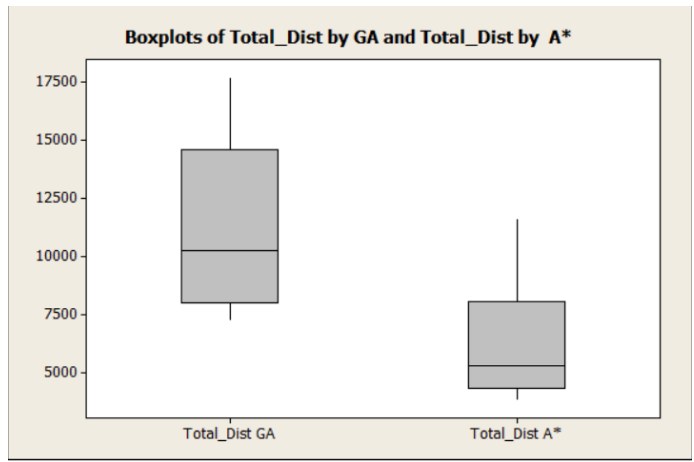

Fig 3: Showing boxplots of total distance using GA and total distance using GA-A*. The distances represented along the y-axis are in meters.

The box plots in Figure 3 clearly show consistently lower values for total distance using the proposed algorithm as compared to the existing GA in ten different routes generated using a single run. Also, the variability in the values of total distances by GA-A* as compared to that of GA is less.

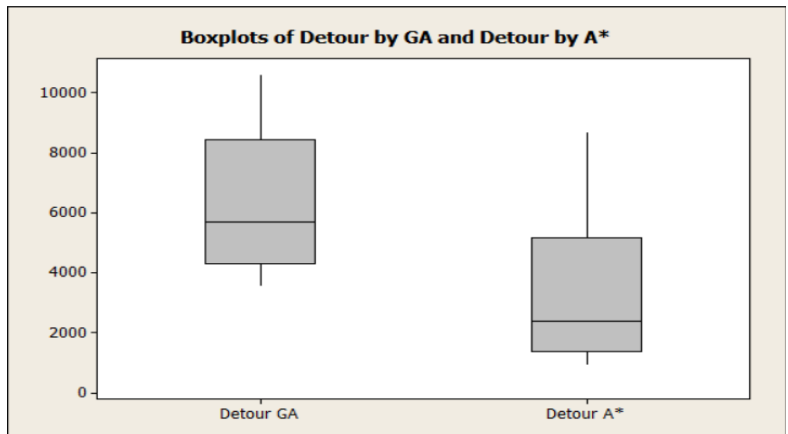

**Fig 4: Showing boxplots of Detour distance using GA and Detour distance using GA-A*. The distances represented along the y-axis are in meters.**

The box plots in Fig 4 clearly show consistently lower values for detour distance using the proposed algorithm as compared to the existing GA in ten different routes generated using a single run.

Also on an average, the total distance and the detour in case of the proposed GA-A* algorithm are clearly less than the existing GA algorithm. The above observations are confirmed using statistical tests given below:

|  | N | Mean | StDev | SE Mean |
|---|---|---|---|---|
| Total_Dist GA | 10 | 11309 | 3629 | 1148 |
| Total_Dist GA-A* | 10 | 6309 | 2618 | 828 |

**Table2: Two-sample t test for Total_Dist GA vs Total_Dist GA-A***

Difference = mean (Total_Dist GA) - mean (Total_Dist GA-A*)
t-Test of difference = 0 (vs >): t-Value = 3.53, P-Value = 0.001

p-value clearly indicates rejection of the null hypothesis that there is no difference in the mean total distances against the greater than type alternative hypothesis at 5% level of significance. Hence we conclude that the average total distance by GA is significantly greater than the average total distance under the proposed algorithm under 5% level of significance.

|  | N | Mean | StDev | SE Mean |
|---|---|---|---|---|
| Detour GA | 10 | 6252 | 2462 | 779 |
| Detour GA-A* | 10 | 3389 | 2618 | 828 |

**Table3: Two-sample t test for Detour GA vs Detour GA-A***

Difference = mean (Detour GA) - mean (Detour GA-A*)
t-Test of difference = 0 (vs >): t-Value = 2.52, P-Value = 0.011

p-value clearly indicates rejection of the null hypothesis that there is no difference in the mean total detour distances against the greater than type alternative hypothesis at 5% level of significance. Hence, we conclude that the average total detour distance by GA is significantly greater than the average total detour distance under the proposed algorithm under 5% level of significance.

The following chart (Figure 5) compares the average Total GA Distance and GA-A* Distance for 20 different executions of the algorithm. It is clear from observation that the average GA-A* distance triumphs over the average distance provided by GA in all executions, thus providing evidence that the routes have been optimized while maintaining an identical level of occupancy.

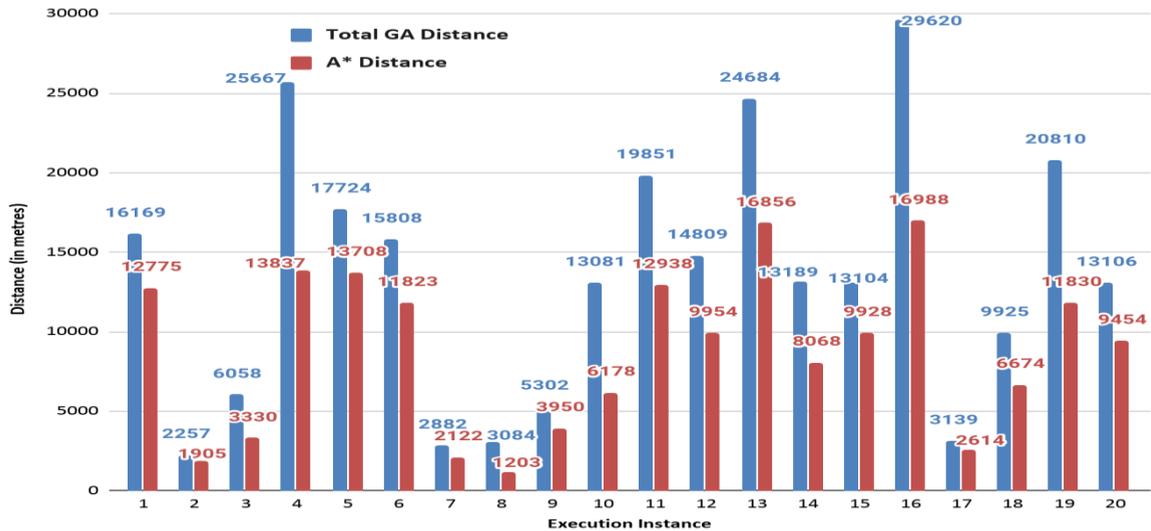

**Fig. 5: Comparison of the average distance per execution of GA and GA-A* for 20 executions.**

The following chart (Figure 6) uses the data from Fig. 5 to demonstrate the percentage improvement in the average distance per execution obtained on using the A* algorithm to optimize the routes generated by the GA.

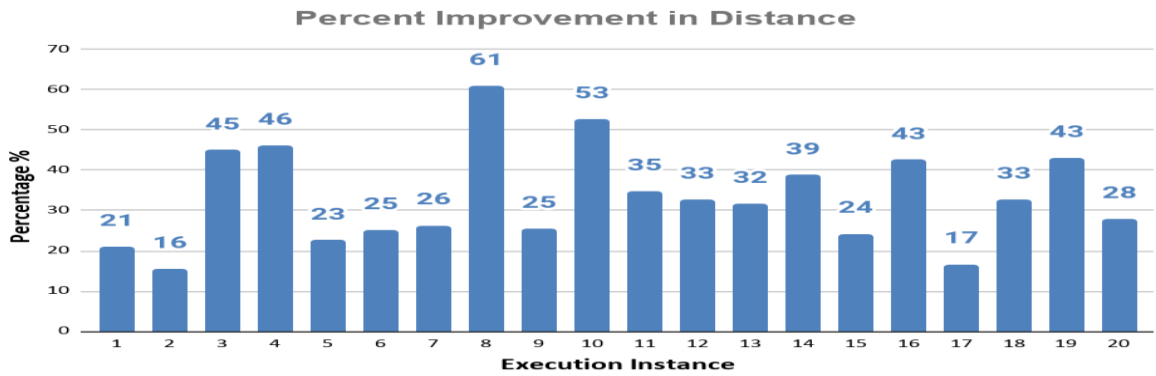
**Fig. 6: Percentage improvement in Distance over 20 executions**

The following three charts compare the distance of the route generated by the GA and the corresponding route generated by GA-A* for three different iterations. It can be observed that every route in each iteration is shortened after optimization using the A* algorithm.

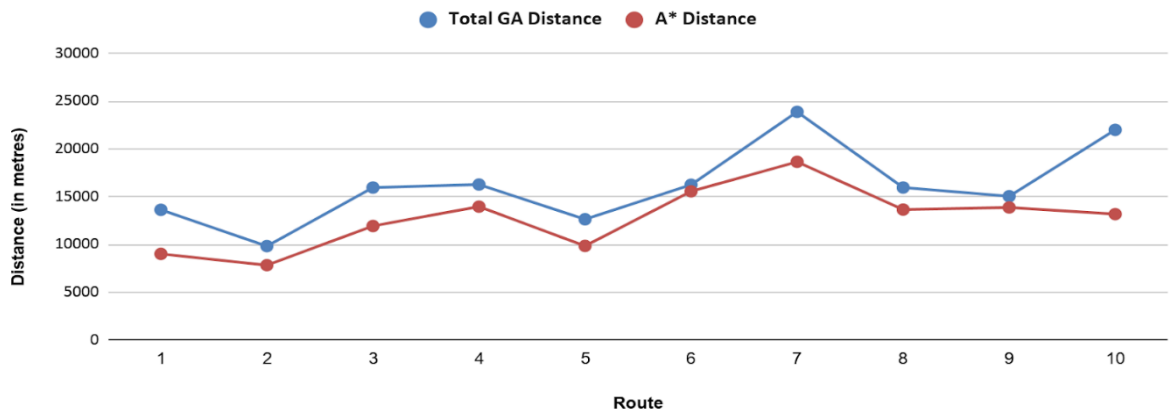

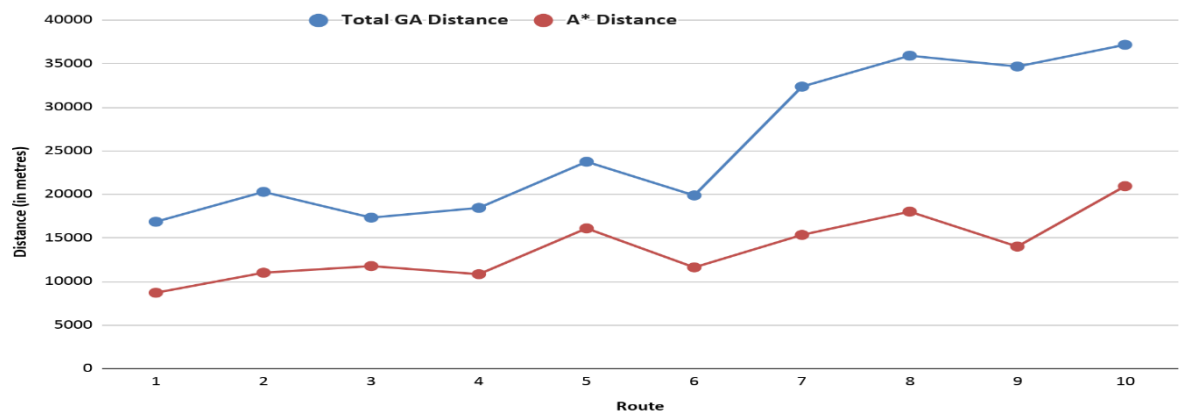

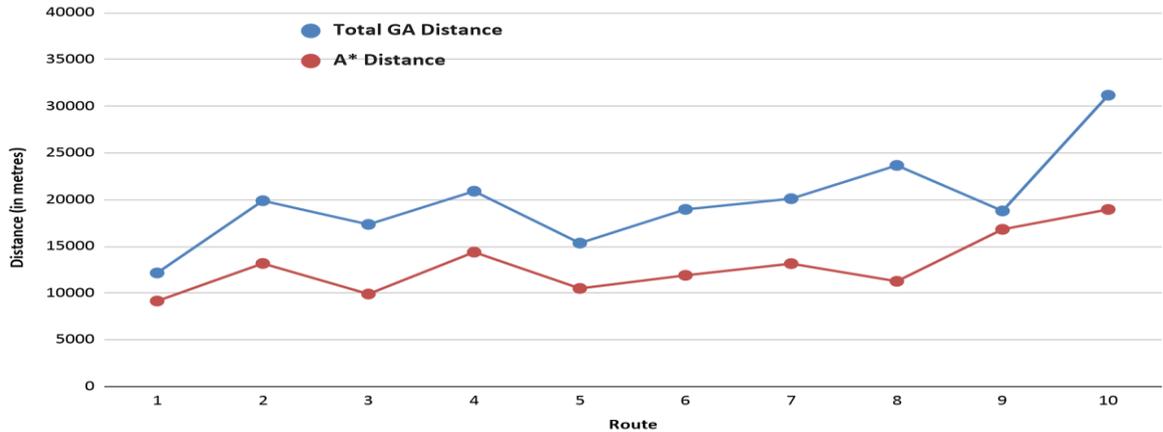

**Fig. 7(a): (a), (b), and (c) compare the distances of the routes generated by GA and GA-A\* in different instances of execution.**

**Statistical Analysis for comparison of Results obtained by Hybrid algorithm and existing algorithm**

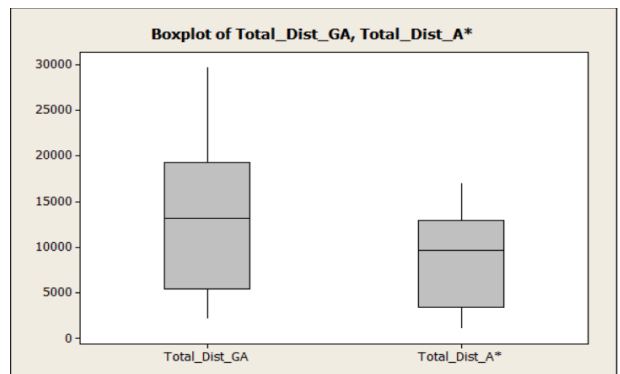

**Fig 8: Boxplots of average total distances generated using GA and that using hybrid GA-A\* for 20 executions. The distances represented along the y-axis are in metres.**

The box plots (Figure 8) are clear indicating lesser average distances by the proposed algorithm as compared to the existing GA algorithm. The proposed algorithm, having lesser spread in the values, seems to be more consistent than the existing one. Statistical test to confirm the above observations are given below:

|  | N | Mean | StDev | SE Mean |
|---|---|---|---|---|
| Total_Dist GA | 20 | 13513 | 8110 | 1813 |
| Total_Dist GA-A* | 20 | 8807 | 5074 | 1135 |

**Table 4: Two-sample t test for average Total_Dist in GA vs average Total_Dist in GA-A* in 20 independent runs**

Difference = mean (Average Total_Dist_GA) - mean(Average Total_Dist_GA-A*)
t-Test of difference = 0 (vs >): t-Value = 2.20 , P-Value = 0.018

p-value clearly indicates rejection of the null hypothesis that there is no difference in the mean total distances against the greater than type alternative hypothesis at 5% level of significance in 20 independent runs of the algorithms. Hence we conclude that the average total distance by GA is significantly greater than the average total distance under the proposed algorithm under 5% level of significance.

## 5. Conclusion

The simulated experiments conducted by this model to provide car-pooling routes in a real road map proved that the proposed Hybrid algorithm, using both GA and A* algorithm, provides more optimal routes than in the case where only GA is used. Also, on an average, the total distance and the detour in case of the proposed algorithm are clearly less than the existing GA algorithm. So, this work has successfully contributed in producing a hybrid algorithm that can be used for real life carpooling situations to provide time efficient results, for both the users and the service provider. From the results it is clear that the routes generated by the existing genetic algorithm are generally longer and the cars have to retract back to the point of diversion to return to the original route after completing the pick-up/drop. However, in real life, it might not be feasible to retract back to the original route. This shortcoming is greatly improved by using the A* algorithm to reroute and optimize the original route. As is evident from the results that have been provided, the optimization provides vast improvements on the previously observed results. This mainly stems from the fact, that A* is an algorithm that emphasizes finding the shortest possible route between two points while limiting the computations to a minimum. The new route that is generated includes the pick-up and drop-off points of the passengers while maintaining the same level of occupancy, thus providing shorter paths. This, in turn, decreases the overall travel of the car, thus ensuring shorter trip times for passengers as well as lesser fuel cost on the part of the driver. This also deals with the negative aspect of having to return to the point of deviation, in cases where it is completely unnecessary. This presents a more practical approach to determining the route thus making the algorithm implementable in the real world.

According to the above experiments and observations, the paths generated by the Genetic Algorithm are not generally the shortest as these paths initially do not include the pick-up and drop-off distance of the passengers. As a result, the cars have to take various detours by deviating from the main route to pick-up or drop-off a passenger and then again return back to the point of deviation. GA generates final paths which are comparatively longer than the paths generated by the optimized GA-A* algorithm. Secondly GA requires considerable longer time to generate the route which is considerably reduced by using A* algorithm. The model proposed by this work can be further improved by taking into consideration various other conflicting objectives like (i) Traffic lights: Larger number of crossings and/or traffic lights along the route may cause congestion and thus loss of time. (ii) Road networks: Aspects like blockage of various roads, or restricted movements of vehicles in particular directions, (iii) Road surface quality: This may also be a factor affecting the choice of routes and can sometimes be very important to reduce the total travel time and (iv) Congestion: There might be an enormous amount of traffic congestion along the shortest route whereas the longer routes may have lesser traffic.